# FRIST: FMRI Representation Informed Shared-space Training Improves EEG-only Individual-Finger BCI Decoding

Jintao Zhang[1, #], Yidan Ding[1, #], Joshua Kosnoff[1], Maxim Karrenbach[2], Hanwen Wang[1], Bin He[1,2,3, *]

[1] Department of Biomedical Engineering, Carnegie Mellon University, Pittsburgh, PA 15213, USA.

[2] Department of Electrical and Computer Engineering, Carnegie Mellon University, Pittsburgh, PA 15213, USA.

[3] Neuroscience Institute, Carnegie Mellon University, Pittsburgh, PA 15213, USA.

# Those authors contributed equally

* Correspondence:

Bin He, PhD

Department of Biomedical Engineering

Carnegie Mellon University

5000 Forbes Avenue, Pittsburgh, PA 15213, USA

**Abstract**

Finger-level motor decoding is important for naturalistic brain-computer interface (BCI) control, yet individual-finger decoding from scalp electroencephalography (EEG) remains challenging because finger representations are spatially close in the sensorimotor cortex and blurred by volume conduction. Leveraging the high spatial resolution of functional MRI (fMRI), we introduce fMRI Representation-Informed Shared-Space Training (FRIST), a two-stage EEG decoding framework that first learns fMRI-informed spectral projections from simultaneous EEG–fMRI recordings and then uses fMRI-derived class geometry to guide residual refinement of EEG predictions. FRIST transfers information across recordings through shared finger labels without requiring paired trials and uses only EEG at inference. We evaluated 12 able-bodied participants during movement execution (ME) and motor imagery (MI) under two-class and three-class chronological session-held-out decoding simulating the online scenario. Using EEGNet as the EEG feature extractor, FRIST increased group average accuracy from 66.93% to 74.53% for two-class ME, from 44.83% to 56.58% for three-class ME, from 80.78% to 85.63% for two-class MI, and from 60.93% to 69.90% for three-class MI compared with the EEG-only EEGNet baseline. FRIST is also shown to improve EEG-only decoding when the target participant's own fMRI data were unavailable. FRIST also generalized across multiple EEG decoding backbones, reaching 87.40% in two-class MI and 72.54% in three-class MI with EEG Conformer as the EEG feature extractor. These findings indicate that fMRI provide useful spatial constraints for EEG representation learning. FRIST improves noninvasive EEG-based finger-level BCI decoding, offering a multimodal strategy for integrating the spatial specificity of fMRI with real-time applicability of EEG.

**Keywords:**

Brain-computer interface, electroencephalography, functional magnetic resonance imaging, motor imagery, individual-finger decoding, supervised contrastive learning, multimodal representation learning.

**Highlights**

- fMRI-informed spectral projections improve EEG-only finger decoding.
- Shared-space training transfers fMRI finger-specific representations to EEG.
- Decoding gains persist without fMRI from the target participant.
- The proposed framework enhances four EEG decoders in finger execution and imagery.

# 1. INTRODUCTION

Brain-computer interfaces (BCIs) translate neural activity into external control signals and have important applications in neurorehabilitation, assistive communication, and restoration of motor function (Ang and Guan, 2015; Ding et al., 2026; Edelman et al., 2025; Pfurtscheller and Neuper, 2001; Wolpaw et al., 2002). Invasive BCIs have demonstrated high-performance control of complex hand and finger movements by recording neural activity directly from cortical or intracortical sites (Hotson et al., 2016; Willsey et al., 2025). These studies show that fine-grained motor information, including individual-finger movement intention, can be extracted from neural activity. However, invasive approaches require implanted electrodes and are constrained by surgical risks, long-term implant stability, and accessibility. Noninvasive BCIs, particularly EEG-based systems, offer a safer and more scalable alternative, although EEG recordings generally provide lower spatial specificity and lower signal-to-noise ratio than invasive recordings.

Among noninvasive BCI paradigms, motor imagery (MI) and movement execution (ME) are especially relevant for rehabilitation and robotic control because they engage sensorimotor networks associated with intended or executed movements (Ang and Guan, 2015; Ding and He, 2026; Edelman et al., 2019; Forenzo et al., 2025, 2024; Meng et al., 2016; Pfurtscheller and Neuper, 2001). Many EEG-based BCI studies focus on coarse motor classes such as hands, feet, or tongue. While these paradigms have enabled robust noninvasive BCI control, more naturalistic and dexterous hand control requires decoding finer motor intentions, including individual-finger movements. Recent noninvasive EEG studies have shown that individual-finger decoding and real-time robotic finger control are feasible (Ding et al., 2025), but robust finger-level decoding, especially for multiple classes, remains challenging because finger-related neural patterns are spatially fine-grained, partially overlapping, and variable across participants.

Electroencephalography (EEG) is widely used for noninvasive BCI because it is low-cost, portable, and provides millisecond-level temporal resolution. Deep learning models such as EEGNet have improved EEG decoding by learning temporal, spectral, and spatial features directly from EEG signals (Lawhern et al., 2018). However, scalp EEG has limited spatial specificity because cortical activity is blurred by volume conduction and mixed at the sensor level (Burle et al., 2015; He et al., 2018). This limitation is particularly problematic for individual-finger decoding, where finger representations are spatially adjacent in the sensorimotor cortex and therefore difficult to separate from scalp recordings.

A complementary neuroimaging modality is functional magnetic resonance imaging (fMRI), which provides high spatial resolution. fMRI has been used to reveal fine-grained finger representations in motor and somatosensory cortices (Diedrichsen et al., 2013; Huber et al., 2020). However, fMRI is expensive, non-portable, and has low temporal resolution, making it unsuitable for routine real-time BCI control (Glover, 2011). These complementary properties suggest a natural multimodal strategy: fMRI may provide spatially specific motor representations to guide EEG feature learning during training, while EEG remains the only signal required during inference.

Prior EEG-fMRI studies have demonstrated that task-related electrophysiological activity and blood-oxygen-level-dependent (BOLD) responses are coupled during ME and MI. Yuan et al.

(2010) reported covariation between sensorimotor EEG rhythms and BOLD responses during motor tasks, indicating that electrophysiological and hemodynamic signals reflect related aspects of motor cortical activity. More recently, Bondi et al. (2025) used EEG-informed fMRI analysis across multiple motor execution and imagery conditions and showed that task-related EEG features are spatially associated with BOLD activation patterns in motor-related brain regions. An important property of this coupling for the present work is that it is expressed at the level of band-limited power. Band power is not a fixed quantity of the EEG but a quantity computed over a chosen spatial direction, so a measured coupling to BOLD can be used not only to interpret EEG activity but also to choose which directions to compute power over. Prior work has used fMRI information to guide EEG analysis in several ways. Task-related fMRI activation has been used to spatially weight EEG features for upper-limb motor classification (Yang et al., 2022), and fMRI-informed time-varying constraints have been incorporated into EEG/MEG source imaging (He and Liu, 2008; Liu and He, 2008; Xu et al., 2018). Other studies have related EEG decoder activity to simultaneously acquired fMRI signals for visualization and closed-loop analysis of sensorimotor activity (Iwama et al., 2024). However, these approaches did not investigate whether fMRI could either select the spatial directions from which EEG features are computed and supply class-specific finger representations as training-time anchors for EEG-only individual-finger decoding. Thus, it remains unclear whether fMRI-derived spatial information can be used to improve EEG-only decoding at the individual-finger level.

Despite the complementarity between EEG and fMRI, fMRI-guided EEG decoding has not been widely developed for practical BCI systems. Simultaneous EEG-fMRI acquisition is technically difficult, affected by scanner-related artifacts, and complicated by the temporal mismatch between electrophysiological and hemodynamic signals (Huster et al., 2012; Warbrick, 2022). In addition, collecting large-scale paired EEG-fMRI datasets is costly. These challenges make direct paired multimodal modeling difficult, especially for finger-level decoding, where the relevant spatial patterns are subtle and difficult to recover from EEG alone. These costs, however, are incurred once. A reference dataset that has already been acquired can in principle be reused indefinitely, if what is learned from it takes a form that applies to new EEG recordings from participants who were never scanned. The methodological challenge is therefore not how to avoid simultaneous EEG-fMRI, but how to spend it once and carry the result forward to EEG-only inference.

Contrastive learning offers a solution for learning shared representations across different modalities by pulling related samples together and pushing unrelated samples apart in a latent space (Chen et al., 2020; Dufumier et al., 2024; Khosla et al., 2020). In BCI research, contrastive learning has been used to improve EEG representation learning under limited labeled data, cross-session variability, and inter-subject variability (Li et al., 2024; Lotey et al., 2022; Zhi et al., 2025). Recent multimodal BCI studies have further extended this idea to cross-modal neural representation learning, including EEG-fNIRS models that align paired electrophysiological and hemodynamic signals in a shared latent space (Jung and An, 2025), as well as self-supervised EEG-fMRI fusion frameworks that synergize complementary information across modalities (Wei et al., 2025). These studies demonstrate the promise of contrastive learning for integrating heterogeneous brain signals. However, existing approaches primarily focused on EEG augmentations, EEG-fNIRS fusion, or general multimodal neuroimaging integration, rather than using high-spatial-resolution fMRI representations as cross-modal anchors for EEG decoding (Kline et al., 2021; Yang et al., 2022). A further constraint applies when the two modalities are acquired in separate sessions, as in the present study, because direct trial-level correspondence is unavailable. Instance-level objectives such as InfoNCE (van den Oord et al., 2019) and CLIP - style contrastive learning (Radford et al., 2021) rely on paired or otherwise matched observations to define positive pairs and therefore cannot be applied directly in this setting. When no trials are shared, the only key common to both modalities is the class label, which makes supervised contrastive learning (Khosla et al., 2020) the appropriate formulation for cross-modal alignment in the absence of shared trials.

Building on this direction, we introduce FRIST (fMRI Representation-Informed Shared-Space Training), a two-stage framework that uses fMRI to guide EEG representation learning while preserving EEG-only inference. In Stage 1, simultaneous EEG–fMRI data are used to estimate frequency-specific spatial projections that identify EEG channel combinations whose band power predicts BOLD activity. These projections are then frozen and applied to out-of-scanner EEG, producing fMRI-informed spectral projection (FISP) features. In Stage 2, block-level fMRI patterns define a fixed, 40-dimensional reference space representing the geometry of the finger classes. A neural network maps the FISP features and frozen EEG-backbone features into an fMRI-aligned EEG representation and learns a residual correction to the Stage 1 class scores. This residual formulation refines an established predictor rather than relearning the complete decision function from scratch(Friedman, 2001; He et al., 2016). The residual output is initialized to zero and trained using complementary classification, residual-direction, fMRI-alignment, and cross-modal contrastive objectives, followed by calibration on the current fold's validation data. Once training is complete, FRIST requires only EEG for decoding. FRIST requires simultaneous EEG-fMRI only once, to fit the projections and to construct the fMRI-derived finger representations. It works without fMRI from the target user and requires no fMRI at inference, so a new user can be decoded without ever entering a scanner. Importantly, FRIST is designed as a flexible framework rather than a stand-alone decoder and can be applied on top of commonly used deep-learning-based EEG feature extractors. Together, these properties allow FRIST to incorporate the spatial specificity of fMRI while retaining the temporal resolution and practical deplorability of EEG.

## 2. METHODS

We developed FRIST (fMRI Representation Informed Shared-space Training) to improve EEG-only individual-finger decoding by incorporating fMRI-derived spatial information into EEG representation learning during training (Fig. 1). The independent fMRI guidance dataset was collected using a block-design ME/MI finger task during simultaneous EEG-fMRI scanning (Fig. 1A), whereas the EEG-only dataset consisted of multiple offline and online EEG sessions evaluated with an online-style session-held-out protocol (Fig. 1B).

FRIST is a two-stage framework that uses fMRI as training-time supervision to improve EEG-only finger decoding (Fig. 1C). Stage 1 derives fMRI-informed spectral projection (FISP) features from simultaneous EEG–fMRI data and combines their predictions with those of a Base EEG model. Stage 2 maps the combined EEG features into a 40-dimensional, fMRI-guided representation and learns a residual correction based on the Stage 1 predictions. After validation-based calibration, all components are fixed and held-out decoding requires only EEG (Fig. 1D).

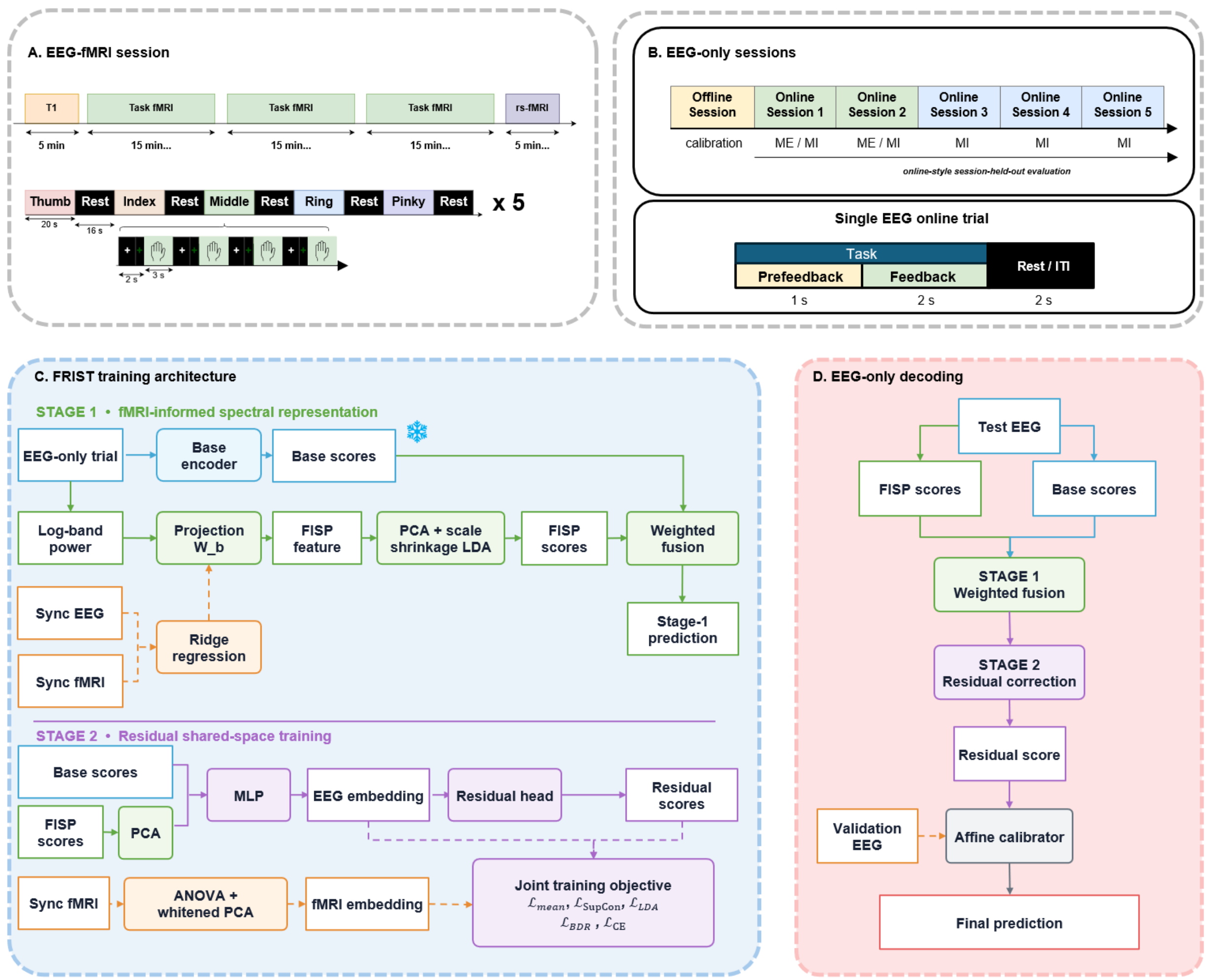

**Fig. 1. Experimental protocol for the simultaneous EEG-fMRI finger ME/MI task (A and B) and the FRIST architecture (C and D).** **A**: Simultaneous EEG-fMRI dataset. Participants completed block-design ME and MI finger tasks during MRI scanning. **B**: EEG-only individual-finger BCI dataset. The EEG dataset included offline/base recordings and multiple online sessions and was evaluated using an online-style session-held-out protocol. **C**: FRIST training involving the fMRI-informed spectral projection in Stage 1 and supervised cross-modal shared space alignment in Stage 2. **D**: During EEG-only inference, held-out EEG passes sequentially through the trained Stage 1 and Stage 2 to produce logits, and the validation-fitted affine calibrator generates the final prediction.

## 2.1. Dataset Description and Task Design

Twelve right-handed able-bodied human subjects (4 male/8 female; mean age: 23.83 ± 3.71) were recruited for this study. All procedures and protocols were approved by the Institutional Review Board of Carnegie Mellon University (protocol number: STUDY2017_00000548). Prior to participation, all subjects completed eligibility screening, were informed of the study procedures and potential risks, and provided written informed consent.

Each participant completed nine EEG-only recording sessions and two simultaneous EEG-fMRI sessions. The EEG-only sessions were used for EEG model training and evaluation, whereas the simultaneous EEG-fMRI sessions were used to obtain fMRI-derived spatial representations for cross-modal guidance.

During the EEG-only sessions, EEG signals were recorded using a 128-channel BioSemi ActiveTwo system (BioSemi, Amsterdam, The Netherlands) at a sampling rate of 1024 Hz. Participants completed three sessions of motor execution (ME) and six sessions of motor imagery (MI) finger movement tasks. For each condition, the first session was conducted offline without feedback, while the remaining sessions provided real-time visual feedback using an EEGNet-based decoder (Lawhern et al., 2018). Each session included both two-class (thumb versus pinky) and three-class (thumb, index, and pinky) decoding tasks. Each online session consisted of 32 runs, with 10 randomized trials per run for each decoding task. During each 3-s trial, participants performed self-paced repetitive flexion and extension of the cued finger using their right hand. For the session-held-out evaluation, two online ME sessions and up to five online MI sessions were reserved for testing, while the sessions before the test session were used for model training and validation. The EEG dataset is publicly available at: https://doi.org/10.1184/R1/29104040.

Each participant contributed nine EEG-only task sessions (three ME and six MI) and two simultaneous EEG–fMRI sessions (one ME and one MI). Simultaneous EEG-fMRI data were collected in two additional sessions after all EEG-only sessions had been completed, one for ME and one for MI, using a block-design paradigm (Fig. 1A). Each task block lasted 20 s and each rest block lasted 16 s. A task block comprised four repetitions of the same finger movement, each lasting 3 s and separated by 2-s inter-trial intervals. Task conditions included five fingers (thumb, index, middle, ring, pinky) with the task sequence randomized at the block level. Participants lay supine with their heads stabilized using foam padding and were instructed to minimize head motion throughout the experiment. Simultaneous EEG was acquired using an MR-compatible BrainAmp MR Plus amplifier (Brain Products GmbH, Gilching, Germany) and a 63-channel BrainCap MR (EasyCap GmbH, Breitbrunn, Germany) arranged according to the international 10–20 system. An additional electrocardiogram (ECG) electrode was placed on the chest for physiological artifact removal. EEG and ECG signals were sampled at 5000 Hz. MRI data were acquired on a 3 T Siemens Magnetom Prisma scanner (Siemens Healthineers, Erlangen, Germany) equipped with a 64-channel head coil at the CMU–Pitt BRIDGE Center (RRID: SCR_023356). Each MRI session included one high-resolution T1-weighted anatomical scan acquired using a three-dimensional magnetization-prepared rapid gradient-echo (MPRAGE) sequence (1 mm isotropic resolution; TR = 2300 ms; TE = 1.9 ms; TI = 900 ms; flip angle = 9°), followed by three task-based functional scans acquired using a gradient-echo echo-planar imaging (EPI) sequence (2 mm isotropic resolution; TR = 2 s; TE = 30 ms; flip angle = 79°; 72 axial slices; 460 volumes per run).

Because simultaneous EEG recordings were limited to a single recording session per condition, they were not used for EEG decoder training. Instead, the simultaneous EEG and fMRI data were used to construct the special-informative representations, whereas all EEG model training, validation, and held-out testing were performed exclusively using EEG-only recordings.

## 2.2. EEG Preprocessing and Input Construction

The EEG-only signals were re-referenced to the common average, downsampled to 100 Hz, and bandpass filtered between 4 and 40 Hz using a fourth-order Butterworth filter. Model-ready trials contained 500 samples for the offline recording and 300 samples for online recordings. Each channel was z-scored within trial and clipped to $[-5,5]$. The normalized trial was segmented into overlapping windows of 100 samples with a stride of 10 samples, corresponding to 1-s windows with 0.1-s strides. A frozen EEG backbone based on EEGNet-8.2 was trained for 40 epochs with learning rate $10^{-4}$. This was served as the Baseline performance. The EEG collected during the simultaneous EEG-fMRI sessions had undergone scanner-gradient and pulse-artifact correction, 0.1–70 Hz filtering with a line-noise notch, and downsampling to 250 Hz. The ECG channel and four peripheral EEG channels (TP9, TP10, FT9, and FT10) were excluded from the present analysis to facilitate mapping between the BioSemi-128 and Brain Product-64 montages, as the BioSemi-128 montage lacked electrodes sufficiently close to these inferior temporal locations. This resulted in a final set of 59 EEG channels.

Then in Stage1, to place the EEG-only recordings in the same channel order as the simultaneous EEG-fMRI recordings, a deterministic Perrin spherical-spline operator mapped 128-channel layout to the 63-channel EasyCap montage (Perrin et al., 1989). Common-average-reference projectors were included on both sides of the geometry-only mapping, after which TP9, TP10, FT9, and FT10 were removed. The mapping was applied to the EEG-only recordings before band-power extraction.

### 2.3 fMRI Preprocessing and Activation Analysis

Raw fMRI volumes were spatially realigned to the first volume to reduce head-motion-related artifacts. Susceptibility-related distortion artifacts were corrected using the topup tool in FMRIB Software Library (FSL; version 6) (Jenkinson et al., 2012). The functional images were then co-registered to each participant's T1-weighted anatomical image, normalized to Montreal Neurological Institute (MNI) standard space, and spatially smoothed using a 3D Gaussian kernel with a full width at half maximum of 4 mm.

Block-design fMRI analyses were performed using custom MATLAB scripts and Statistical Parametric Mapping software (SPM12) to localize task-related hemodynamic responses during the scanner task. For each participant included in the group analysis, first-level contrast images were generated for each finger-versus-rest comparison separately for ME and MI, resulting in ten contrast images per participant: five fingers for ME and five fingers for MI. These first-level contrast maps were entered into second-level random-effects analyses, with age, sex, and handedness included as covariates. Group-level $t$-contrasts were applied to evaluate task-related responses.

For the ME-versus-MI second-level comparison, a repeated-measures flexible factorial model was implemented in SPM12 with three factors: Subject, Task, and Finger. Subject was modeled to account for inter-individual variability, whereas Task and Finger were modeled as repeated-measures factors. $T$-contrasts were specified to test the main task effect of ME > MI. Reported statistical maps were thresholded at a voxel-level threshold of $p < 0.001$ uncorrected, followed by cluster-level family-wise error correction at $p_{\mathrm{FWE}} < 0.05$. Significant clusters were anatomically labeled using AAL3 (Rolls et al., 2020).

The spatial organization of finger-related activity for ME and MI was quantified by computing the weighted average of all suprathreshold voxel coordinates according to the positive finger-versus-rest thresholded $T$ maps. The resulting weighted centroid coordinates were visualized on a top-view glass-brain projection to show the relative spatial arrangement of the five finger representations. To assess the similarity of finger-specific activation patterns, unthresholded $T$ maps for the five finger-versus-rest contrasts were masked by the union of the corresponding thresholded finger activation regions. Within this mask, each unthresholded finger $T$ map was z-scored. Pairwise Pearson correlations were then computed between the voxelwise $t$-statistics for all finger pairs for ME and MI tasks separately.

## 2.4. fMRI Representation Construction

Stage 1 used temporally aligned, volume-level EEG band-power and BOLD observations acquired during simultaneous EEG–fMRI to estimate band-specific EEG-to-BOLD projections. In contrast, Stage 2 used block-level fMRI activation patterns to define the representational geometry of the finger classes.

For Stage 2, each voxel's BOLD time series was z-scored within run. Event sequences were convolved with the canonical SPM hemodynamic response function, and the volumes associated with each response interval were averaged to obtain a single block-level activation pattern. Voxels from bilateral primary somatosensory cortex (S1), primary motor cortex (M1), and primary visual cortex (V1) according to AAL3 labeling were concatenated, yielding 22,154 voxels per block. Block-level patterns were then standardized separately within each participant. A univariate one-way analysis of variance (ANOVA) was used to retain the 4,000 voxels with the largest F statistics for discrimination among the decoded finger classes. The selected features were standardized, and whitened principal component analysis (PCA) was applied to obtain a fixed 40-dimensional fMRI representation. Separate fMRI reference representations were constructed for ME and MI, as well as for the two-class and three-class decoding settings.

In the full-cohort condition, the Stage 1 projections and Stage 2 fMRI reference representations were estimated using simultaneous EEG–fMRI data from all 12 participants. In the participant-excluded condition, both components were re-estimated separately for each target participant using data exclusively from the remaining 11 participants.

## 2.5. fMRI Representation-Informed Shared-Space Training

### 2.5.1. Stage 1: fMRI informed spectral representation

EEG recorded in the simultaneous EEG-fMRI session was decomposed into five frequency bands: theta (4–7 Hz), alpha (8–12 Hz), low beta (13–20 Hz), high beta (21–30 Hz), and low gamma (31–40 Hz). Each band was obtained with a fourth-order zero-phase Butterworth filter and Hilbert transform. Power was averaged in 100-ms bins, convolved with the canonical hemodynamic response function, and averaged within each fMRI volume. Scanner EEG power and S1–M1–V1 BOLD signals were standardized within participant; finger means and linear and quadratic block-position effects were removed from both modalities. The residual BOLD target was standardized and reduced to 60 principal components.

For every frequency band, ridge regression with the same fixed penalty, $\alpha=1000$, mapped the standardized 59-channel EEG residuals to the 60-dimensional BOLD target (Hoerl and Kennard, 1970). The leading 30 right singular vectors of the fitted coefficient matrix were converted back to the original channel units and stored as a $30\times59$ projection. The same fixed ridge formulation and penalty were used for every frequency band, task, class setting and reference condition.

Each 3-s EEG-only trial was mapped to the 59-channel EasyCap montage and processed using the same five log-power bands and 30 nonoverlapping 100-ms bins. Applying the frozen $30\times59$ projection to each band-specific channel-by-time power map yielded $5\times30\times30=4{,}500$ FISP features per trial.

Within each chronological EEG fold, PCA was fitted to the training-set FISP features, and up to six geometrically spaced component counts were compared on the current validation set. For each count, PCA scores were standardized using training-set statistics and classified with automatic-shrinkage linear discriminant analysis (LDA) (Ledoit and Wolf, 2004). The configuration with the highest validation accuracy was then fixed for held-out testing.

The Base (frozen EEGNet) and FISP classifiers produced one class-score vector per trial. Before fusion, each score vector was centered by subtracting its mean across classes. The two branches were then independently calibrated by temperature scaling, with temperatures estimated on the target participant's current-fold validation set by minimizing negative log-likelihood (Guo et al., 2017). For trial $i$, the Stage 1 score vector was calculated as

$$\mathbf{s}_i^{(1)} = \mathcal{C}\left[\frac{w}{T_{\mathrm{B}}}\,\mathcal{C}(\mathbf{b}_i) + \frac{1-w}{T_{\mathrm{F}}}\,\mathcal{C}(\mathbf{f}_i)\right] \tag{1}$$

Here, $\mathbf{b}_i$ and $\mathbf{f}_i$ are the Base and FISP class-score vectors, and $\mathbf{s}_i^{(1)}$ is the resulting Stage-1 score vector; all three belong to $\mathbb{R}^K$. $K$ is the number of decoded classes; $T_{\mathrm{B}}$ and $T_{\mathrm{F}}$ are positive validation-fitted temperatures; and $w$ is the Base weight. The operator $\mathcal{C}(\mathbf{v}) = \mathbf{v} - \frac{\mathbf{1}_K{}^{\mathrm{T}}\mathbf{v}}{K}\,\mathbf{1}_K$ subtracts the within-trial mean across classes. The weight selection was based on mean validation accuracy over validation folds from the other 11 participants together with the target participant's current-fold validation data. The selected temperatures and weight were fixed during evaluation.

Static5 served as an fMRI-free spectral control to test whether improvements from FISP reflected the learned EEG–BOLD projection rather than the generic benefit of adding a second EEG-only band-power classifier. For each EEG-only trial, it used one whole-trial log-power value from each of the five frequency bands and each of the 128 BioSemi channels. These features underwent the same training-only PCA, standardization, shrinkage-LDA classification, temperature calibration, and two-branch fusion with the Base classifier used for FISP. Static5 did not use the 59-channel montage, simultaneous EEG–fMRI data, or any EEG-to-BOLD projection.

### 2.5.2. Stage 2: Residual shared-space training

Stage 2 jointly learned a residual correction to the Stage 1 class scores and a 40-dimensional EEG representation aligned with fMRI-derived finger-class geometry through class-mean, LDA-based, and supervised contrastive objectives.

#### 2.5.2.1. EEG and fMRI Inputs

Within each EEG fold, the frozen 96-dimensional Base feature was standardized with training-set statistics. The 4,500-dimensional FISP vector was separately reduced to 128 principal components using the training trials only and rescaled by one training-derived scalar so that its total variance matched that of the backbone block. Concatenation therefore produced a 224-dimensional EEG input. The fixed fMRI input was the corresponding 40-dimensional representation described above.

#### 2.5.2.2. Residual shared-space network

The EEG input passed through a linear 224→256 layer, GELU, layer normalization, dropout 0.3, a linear 256→40 layer, and nonaffine layer normalization. A zero-initialized bias-free linear head mapped the 40-dimensional EEG embedding to K residual scores. During training, the raw residual was tanh-bounded, centered across classes, and weighted by the normalized entropy of the Stage-1 prediction:

$$\tilde{\mathbf{s}}_i = \mathcal{C}\left(\mathbf{s}_i^{(1)}\right) + g_i \mathcal{C}\left[\tanh\left(\mathbf{q}_i\right)\right]; \; g_i = \frac{-\sum_{k=1}^{K} p_{ik} \log\left(p_{ik}\right)}{\log K}, \; \mathbf{p}_i = \text{softmax}\left[\mathcal{C}\left(\mathbf{s}_i^{(1)}\right)\right] \quad (2)$$

Here, $\mathbf{q}_i$ and $\tilde{\mathbf{s}}_i \in \mathbb{R}^K$ are the raw residual-score vector and the corrected score vector used during Stage-2 training, respectively. The vector $\mathbf{p}_i \in [0,1]^K$, with $\sum_{k=1}^{K} p_{ik} = 1$, is obtained by applying softmax to the centered Stage-1 scores. $p_{ik}$ denotes its $k$th component. $g_i \in [0,1]$ is its normalized entropy; $\log$ denotes the natural logarithm, $k$ indexes finger classes, and $\tanh$ is applied elementwise. Thus, uncertain trials receive a larger residual correction. Because the residual head is initialized at zero, $\mathbf{q}_i = \mathbf{0}$ initially and $\tilde{\mathbf{s}}_i = \mathcal{C}\left(\mathbf{s}_i^{(1)}\right) = \mathbf{s}_i^{(1)}$, exactly recovering Stage 1. During optimization, $\mathbf{q}_i$ denotes the output under the current network parameters.

### 2.5.2.3. Training Objective

Training combined five complementary loss terms. First, class-balanced cross-entropy optimized the Stage 1 scores after applying the Stage 2 residual correction. Second, a bounded residual-direction loss encouraged this correction to match the difference between the true class label and the Stage 1 probability distribution. Two fMRI-guided losses aligned the EEG and fMRI class means and required the EEG embeddings to be correctly classified by an LDA boundary fitted exclusively to the fMRI reference bank. Finally, a cross-modal supervised contrastive loss pulled each EEG embedding toward same-finger fMRI representations and away from other-finger representations (Khosla et al., 2020). The total normalized loss was

$$\mathcal{L} = \frac{\mathcal{L}_{\text{CE}}}{c_{\text{CE}}} + 0.5 \frac{\mathcal{L}_{\text{BDR}}}{c_{\text{BDR}}} + 0.25 \frac{\mathcal{L}_{\text{mean}}}{c_{\text{mean}}} + 0.25 \frac{\mathcal{L}_{\text{LDA}}}{c_{\text{LDA}}} + 0.1 \frac{\mathcal{L}_{\text{SC}}}{c_{\text{SC}}} \quad (3)$$

Here, $\mathcal{L}$ is the total Stage-2 training objective, $\mathcal{L}_{\text{CE}}$ is class-balanced cross-entropy on the corrected EEG scores $\tilde{\mathbf{s}}_i$; $\mathcal{L}_{\text{BDR}}$ is the mean-squared difference between the centered bounded residual and the correction target formed by the one-hot label minus the Stage-1 class-probability vector. This target is the negative gradient of multiclass cross-entropy with respect to the Stage-1 logits and therefore supplies a stagewise residual-fitting direction (Friedman, 2001); $\mathcal{L}_{\text{mean}}$ is the class-balanced mean-squared distance between each EEG class mean and the corresponding fixed fMRI class mean; $\mathcal{L}_{\text{LDA}}$ is class-balanced cross-entropy after the EEG embedding is scored by the fixed LDA weights and intercept fitted exclusively to the fMRI representations; and $\mathcal{L}_{\text{SC}}$ is the class-balanced mean of the one-directional EEG-anchor-to-fMRI supervised contrastive losses. Before similarity computation, EEG embeddings and layer-normalized fMRI-bank rows were $\ell_2$-normalized. Same-finger fMRI rows were positive, other-finger rows were negative, and the contrastive temperature was $\tau = 0.1$. For each loss term $\mathcal{L}_\ell$, the positive scalar $c_\ell$ was set from its initial loss value before the first parameter update and then held fixed throughout Stage 2 training. The lower bound was 0.05 for the bounded-residual loss and $10^{-8}$ for all other losses. The five coefficients in Eq. (3): 1, 0.5, 0.25, 0.25, and 0.1, were used in every task, reference condition, and backbone.

The network was trained independently within each EEG fold for 400 epochs using AdamW with a learning rate of $10^{-3}$ and weight decay of $10^{-4}$ (Loshchilov and Hutter, 2018). To improve model stability, parameters from every update between 100 and 400 were accumulated into an online arithmetic average, which defined the final fitted EEG encoder and residual head.

#### 2.5.2.4. Validation-only Residual Calibration

An affine calibration layer was fitted on the validation set for each fold to combine the Stage 1 scores with the Stage 2 residual corrections.

$$\mathbf{s}_i^{(2)} = \mathbf{a} \odot \mathbf{s}_i^{(1)} + \mathbf{B}\,\mathcal{C}\left[\tanh\left(\bar{\mathbf{q}}_i\right)\right] + \boldsymbol{\beta} \quad (4)$$

In Eq. (4), $\mathbf{s}_i^{(2)} \in \mathbb{R}^K$ is the final calibrated score vector; $\mathbf{s}_i^{(1)} \in \mathbb{R}^K$ is the Stage-1 score vector; and $\bar{\mathbf{q}}_i \in \mathbb{R}^K$ is the raw residual-score vector produced by the parameter-averaged Stage-2 network. The vector $\mathbf{a} \in \mathbb{R}^K$ contains per-class Stage-1 scales, $\mathbf{B} \in \mathbb{R}^{K\times K}$ mixes the centered bounded residual scores, $\boldsymbol{\beta} \in \mathbb{R}^K$ is the calibration bias, and $\odot$ denotes elementwise multiplication. $\mathcal{C}$ is the class-centering operator defined in Eq. (1). The entropy gate in Eq. (2) was used only to train the Stage-2 network; the validation calibrator in Eq. (4) acted directly on the centered, bounded output $\mathcal{C}[\tanh(\bar{\mathbf{q}}_i)]$. The calibrator was fitted for nine Adam updates using only the validation set, with cross-entropy plus 0.77 times a hinge penalty that penalized a correct-class margin below 0.5, plus 0.006 times an identity regularizer on the Stage-1 scales, residual-mixing matrix, and bias. All fitted quantities were fixed before test inference.

At inference, a held-out EEG trial passes through the frozen Base and FISP paths, the trained EEG projector and residual head, and the validation-fitted calibrator. The resulting calibrated logits constituted the final evaluation scores, with the highest-scoring class selected as the predicted finger.

### 2.6. Evaluation Protocols

Finger discriminability reflected in the fMRI data was evaluated using three-fold within-participant cross-validation. For each participant and task condition, one scanner run was held out in each fold, and the remaining runs were used for model fitting. Flattened BOLD signals within S1, M1, and V1 were classified by logistic regression under the ME and MI two- and three-class settings.

For the main EEG decoding experiments, we used a chronological session-held-out protocol to approximate prospective session-to-session decoding. For each participant, each eligible online session served once as the held-out test set. Only data acquired before the test session, including the participant's preceding offline and online sessions, were available for model development. Within this historical training pool, 20% of the trials were randomly assigned to the validation set, and the remaining trials were used for training.

To assess sensitivity to inclusion of the target participant's reference recording, we repeated the analysis under a participant-excluded reference condition. Both the Stage 1 scanner EEG–fMRI projection bank and the Stage 2 fMRI representation bank were rebuilt using only the other 11 participants. The target participant's EEG-only training, validation, and test partitions were unchanged. For the direct comparison between the all-participant and participant-excluded reference conditions, both conditions were evaluated using the same strict fold-local, validation-only fusion protocol.

Performance across backbones was evaluated by replacing EEGNet with EEG Conformer (Song et al., 2023), a residual dilated temporal convolutional network (TCN) (Bai et al., 2018), or an 1D-CNN-Transformer from Peng et al. (2025). Each backbone was trained for 40 epochs at learning rate $10^{-4}$. The trained backbone was then frozen, and all other FRIST settings were unchanged.

### 2.7. Representation Analysis Metrics

The correct-centroid margin tested whether a held-out EEG embedding was closer to the fMRI centroid for its true finger than to the competing fMRI centroids.

$$M_i = \cos\left(\mathbf{z}_i, \boldsymbol{\mu}_{y_i}\right) - \frac{1}{K-1} \sum_{k \neq y_i} \cos\left(\mathbf{z}_i, \boldsymbol{\mu}_k\right) \tag{5}$$

Here, $M_i \in \mathbb{R}$ is the trial-level correct-centroid margin; $\mathbf{z}_i \in \mathbb{R}^{40}$ is the shared EEG embedding for held-out trial $i$; $y_i \in \{1, \dots, K\}$ is its true class. Each fixed fMRI-bank row was first nonaffinely layer-normalized. Rows were averaged within class, after which the resulting class mean was $\ell_2$-normalized; $\boldsymbol{\mu}_k \in \mathbb{R}^{40}$ is the $\ell_2$-normalized prototype for class $k$. $K$ is the number of classes, $k$ indexes classes, and the sum in Eq. (5) runs over $k \in \{1, \dots, K\} \setminus \{y_i\}$. The held-out EEG embedding was likewise $\ell_2$-normalized before comparison. The function $\cos(\cdot,\cdot)$ denotes cosine similarity. A positive value indicates greater similarity to the matching fMRI centroid than to the mean of the competing centroids. Within each held-out session, trial margins were first averaged within class and then averaged equally across the $K$ classes; the resulting session-level margins were subsequently averaged within participant.

### 2.8. Statistical Analysis

For the group-level EEG decoding and representation analyses, participants were the statistical units ($n = 12$). Group results are reported as the mean across participants $\pm$ sample standard deviation, and accuracy changes are expressed as absolute percentage-point differences. Paired differences were evaluated using two-sided $t$ tests with Cohen's $d_z$, whereas prototype margins were evaluated using two-sided one-sample $t$ tests against zero. Benjamini–Hochberg false-discovery-rate correction (Benjamini and Hochberg, 1995) was applied for multiple-comparison corrections. Adjusted p values, denoted $p_{\mathrm{FDR}}$, were used for statistical interpretation. For the group-level fMRI activation maps, a voxel-level threshold of $p < 0.001$, uncorrected, was followed by cluster-level family-wise-error correction at $p_{FWE} < 0.05$.

## 3. RESULTS

In this study, we first assessed whether fMRI activation patterns contained finger-specific spatial information that could provide meaningful supervisory guidance for EEG representation learning. We then evaluated whether the proposed fMRI Representation Informed Shared-space Training framework improved EEG-only individual-finger decoding relative to EEGNet-based baselines. Finally, we tested the scalability and generalizability of the framework by evaluating decoding performance when the target participant's fMRI data were excluded from the fMRI feature bank and by applying the approach across multiple deep learning EEG decoders.

### 3.1. fMRI Reveals Finger-Specific Spatial Representations

Group-level fMRI analysis showed finger-related activation in motor-related cortical and subcortical regions. Across individual finger movements, BOLD activation showed substantial overlap within a distributed sensorimotor network. For motor execution, all five fingers showed positive activation involving the left primary motor cortex, supplementary motor area (SMA), and basal ganglia, particularly the lenticular nucleus/putamen (Fig. 2A, Supplementary Table S1). These shared regions suggest that individual finger movements recruit a common contralateral sensorimotor and motor-planning network. Motor imagery also elicited overlapping activation across fingers, but the shared network appeared more spatially restricted than that observed during motor execution (Fig. 2B, Supplementary Table S2). The supplementary motor area was the most consistent region across the five finger imagery contrasts, with bilateral involvement observed across fingers.

In contrast, primary sensorimotor regions were less consistently present across imagery contrasts than execution contrasts, appearing most clearly for the middle and pinky finger imagery conditions. Together, these findings indicate that the execution of different individual fingers is supported by a largely shared motor network, with finger-specific differences likely reflected in the spatial distribution and relative strength of activation within this network. In contrast, finger motor imagery recruits a core motor-planning network shared across fingers, while producing weaker or less spatially extensive activation in primary sensorimotor cortex compared with overt finger execution.

Although the individual finger neural activation patterns revealed a common sensorimotor network, the activation maps also suggest potential finger-dependent spatial patterns. During motor execution, peaks in the left precentral and postcentral gyri were consistently located in the hand area, but their peak coordinates varied across fingers, suggesting that finger identity may be represented by partially shifted or partially overlapping activation patterns within the sensorimotor cortex (supplementary Table S1). The centroid of the finger activation weighted by the t-statistics indicated a lateral-to-medial gradient pattern from thumb to pinky (Fig. 2C), while the differentiation for MI activities is smaller (Supplementary Fig. S1A). Neighboring fingers share more similar activation patterns compared to fingers that are farther away (Fig. 2D, Supplementary Fig. S1B).

Comparing motor execution and motor imagery, the current results suggest that execution produced stronger and more extensive activation than imagery (Fig. 2E). Motor execution showed more positive clusters and broader recruitment, primarily in the right cerebellum and left sensorimotor cortex. Motor imagery, by contrast, was dominated by SMA, with fewer clusters overall. This pattern is consistent with the interpretation that motor execution engages both motor planning and overt sensorimotor output pathways, whereas motor imagery preferentially engages internal motor planning networks with reduced activation of primary sensorimotor and cerebellar execution-related circuits.

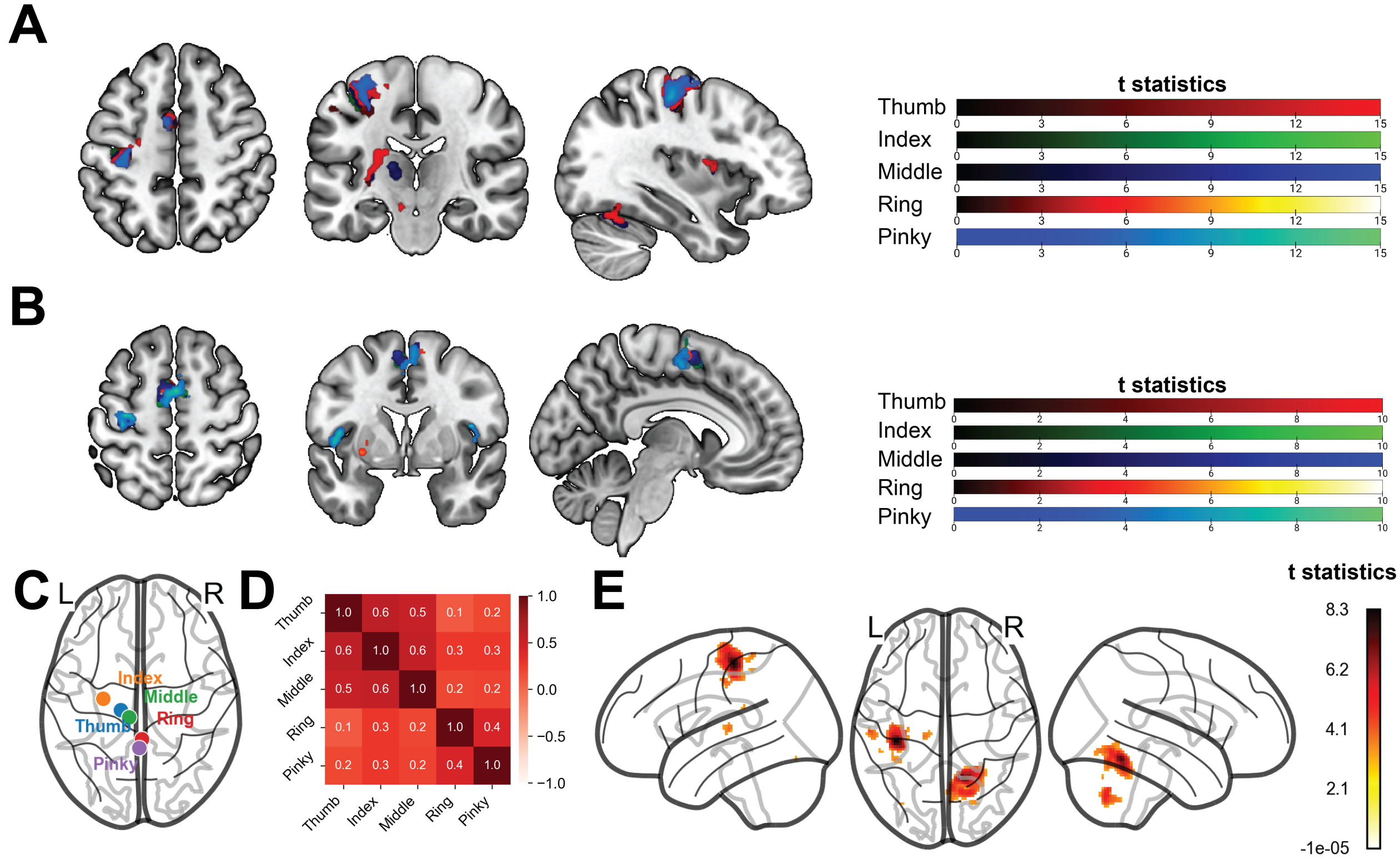

**Fig. 2. Group-level fMRI activation during individual finger movement and imagery. A:** Group-level fMRI statistical maps for finger ME conditions, showing the BOLD activations during individual finger tasks ($p < 0.05$, cluster-level family-wise error corrected) (n = 12) [x = −35, y = −20, z = 51]. **B:** Group-level fMRI statistical maps for finger MI conditions, showing the BOLD activations during individual finger tasks ($p < 0.05$, cluster-level family-wise error corrected) (n = 12) [x = -6, y = 1, z = 58]. **C**: T-statistic-weighted centroids for individual finger activation maps during ME. **D**: Pairwise Pearson correlations between voxel-wise spatial activation patterns for individual fingers during ME. Cell values indicate correlation coefficients, with greater values denoting greater spatial similarity. **E**: Group-level fMRI statistical maps showing the BOLD activations during individual finger tasks that are significantly stronger in ME tasks compared to MI tasks ($p < 0.05$, cluster-level family-wise error corrected) (n = 12).

**TABLE I.** Modality-level decoding performance. Values are test accuracy from within-subject 3-fold cross-validation decoding for fMRI and EEG, reported as mean ± SD across all 12 participants.

| Model | ME 2-class Accuracy | ME 3-class Accuracy | MI 2-class Accuracy | MI 3-class Accuracy |
|---|---|---|---|---|
| fMRI-only Logistic Regression | 92.50±8.78% | 82.96±14.54% | 80.56±15.56% | 70.00±16.51% |
| EEG-only EEGNet | 65.28±8.76% | 41.30±6.83% | 70.00±12.84% | 46.67±8.44% |

We next characterized whether the fMRI guidance data contained finger-discriminative information that could serve as spatial supervision for EEG representation learning. Compared with the EEGNet offline baseline, fMRI decoding showed substantially higher accuracies, especially in the three-class settings. This supports the use of fMRI as a spatially informative source of supervision for EEG representation learning. Across all 12 participants, fMRI-only accuracy was 92.50% and 82.96% for two- and three-class ME and 80.56% and 70.00% for the corresponding MI settings (Table I). These values indicate that the fMRI guidance bank contained strong finger-discriminative information across both execution and imagery conditions. Consistent with the group-level activation analysis, the ME fMRI data were more discriminative than the MI data.

### 3.2. FRIST Improves EEG-Only Individual-Finger Decoding Through Two Stages

Across all four decoding conditions, the FRIST framework significantly outperformed the matched EEGNet Base model (Fig. 3). For motor execution (ME), FRIST increased the two-class accuracy fromTwo-class ME accuracy increased from 66.93±7.06% to 74.53±8.83% ($p_{\mathrm{FDR}}$=0.0015), and three-class ME increased accuracy from 44.83±5.27% to 56.58±9.39% ($p_{\mathrm{FDR}}$=0.0002). For motor imagery (MI), the two-class accuracy increased from Two-class MI increased from 80.78±6.08% to 85.63±5.15% ($p_{\mathrm{FDR}}$=0.0003), whereas the three-class accuracy increased fromand three-class MI increased from 60.93±5.88% to 69.90±4.57% ($p_{\mathrm{FDR}}$<0.0001).

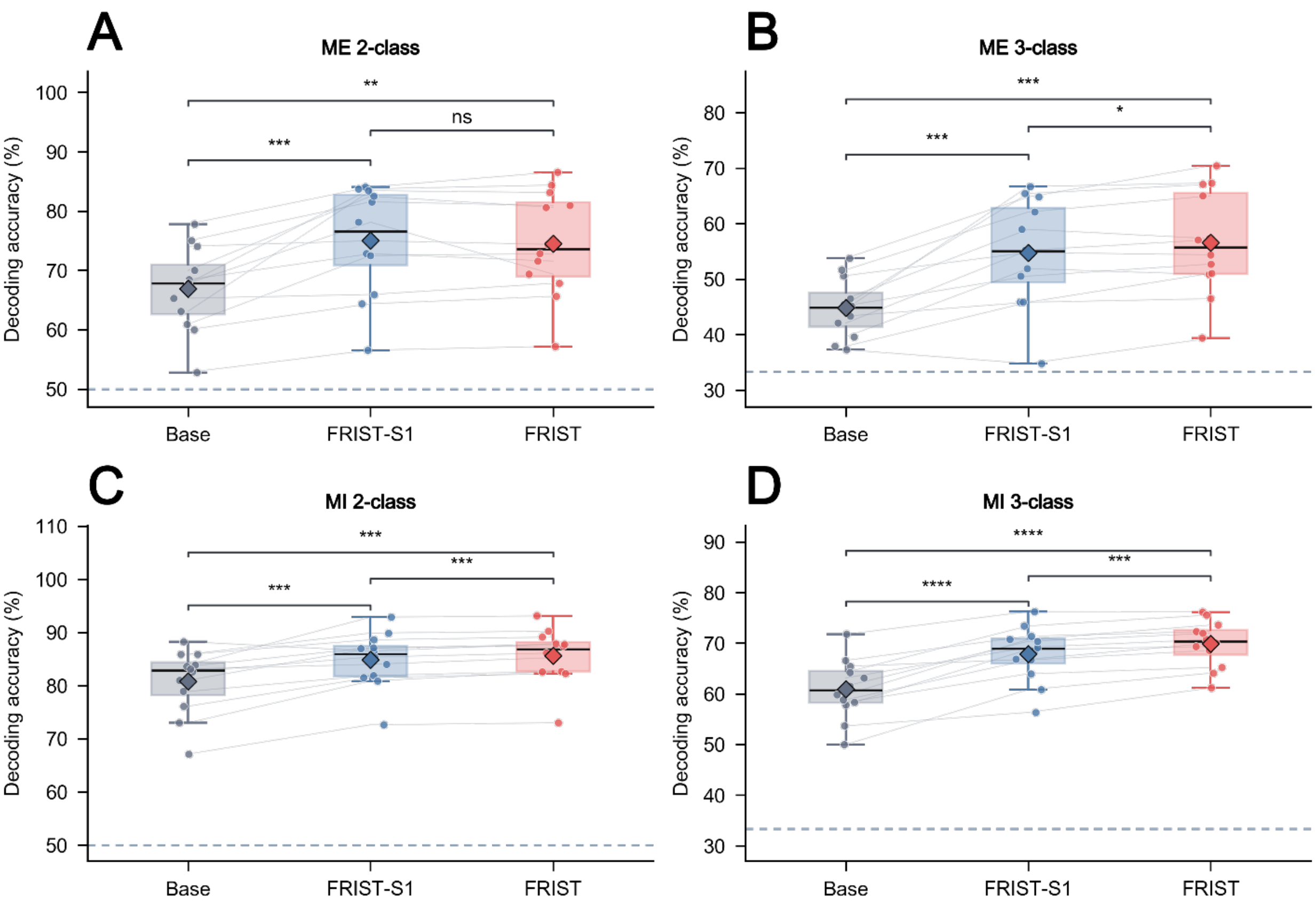

**Fig. 3. Participant-level decoding accuracy across the three decoders. A**: Two-class ME. **B**: Three-class ME. **C**: Two-class MI. **D**: Three-class MI. Each panel compares Base, FRIST-S1 (Stage-1), and FRIST. Boxes show interquartile ranges, center lines indicate medians, diamonds denote means, and points show participants. Gray lines link observations from the same participant, and dashed lines indicate chance. Brackets report two-sided paired tests after Benjamini–Hochberg correction (*, $p_{FDR} < 0.05$; **, $<0.01$; ***, $<0.001$; ****, $<0.0001$; ns, no significance).

Analysis of the nested model configurations further distinguished the contributions of the two stages. In Stage 1, FRIST-S1 significantly improved accuracy over the Base model by 8.12, 9.91, 4.07, and 6.96 percentage points for two-class ME, three-class ME, two-class MI, and three-class MI, respectively. Thus, the BOLD-guided spectral representation provided a significant performance benefit in every decoding condition.

In Stage 2, incorporating the residual shared-space and contrastive learning further provided significant additional gains in three-class ME ($\Delta = 1.85\%, p_{\mathrm{FDR}} = 0.0237$), two-class MI ($\Delta = 0.77\%$, $p_{\mathrm{FDR}} = .0003$), and three-class MI ($\Delta = 2.00\%, p_{\mathrm{FDR}} = .0006$) conditions, but not in two-class ME. Collectively, these results indicate that the BOLD-guided spectral representation accounted for the primary improvement, while the shared-space contrastive learning provided complementary discriminative information in three of the four decoding conditions. Complete summaries and paired statistics for the three nested systems are provided in Supplementary Table S3.

### 3.3. fMRI-Informed Spectral and Cross-Modal Representation Analyses

To assess whether the Stage-1 improvement was attributable to the fMRI-guided spectral representation rather than merely to expanding the EEG representation across five frequency bands, we compared FRIST-S1 with Static5, an EEG-only spectral control approach (Fig. 4A; Supplementary Table S4). Static5 extracted whole-trial log-power features from the same five bands and used the same downstream dimensionality-reduction, classification, and fusion procedures as in FRIST-S1, but without BOLD-guided direction selection. FRIST-S1 outperformed Static5 by 5.49, 6.82, 4.39, and 4.67 percentage points for two-class ME, three-class ME, two-class MI, and three-class MI, respectively, and all four comparisons remained significant after multiple-comparison correction. The consistent improvement indicates that the Stage-1 gain cannot be explained solely by the inclusion of five-band power features and supports the added value of fMRI guidance in constructing the spectral representation.

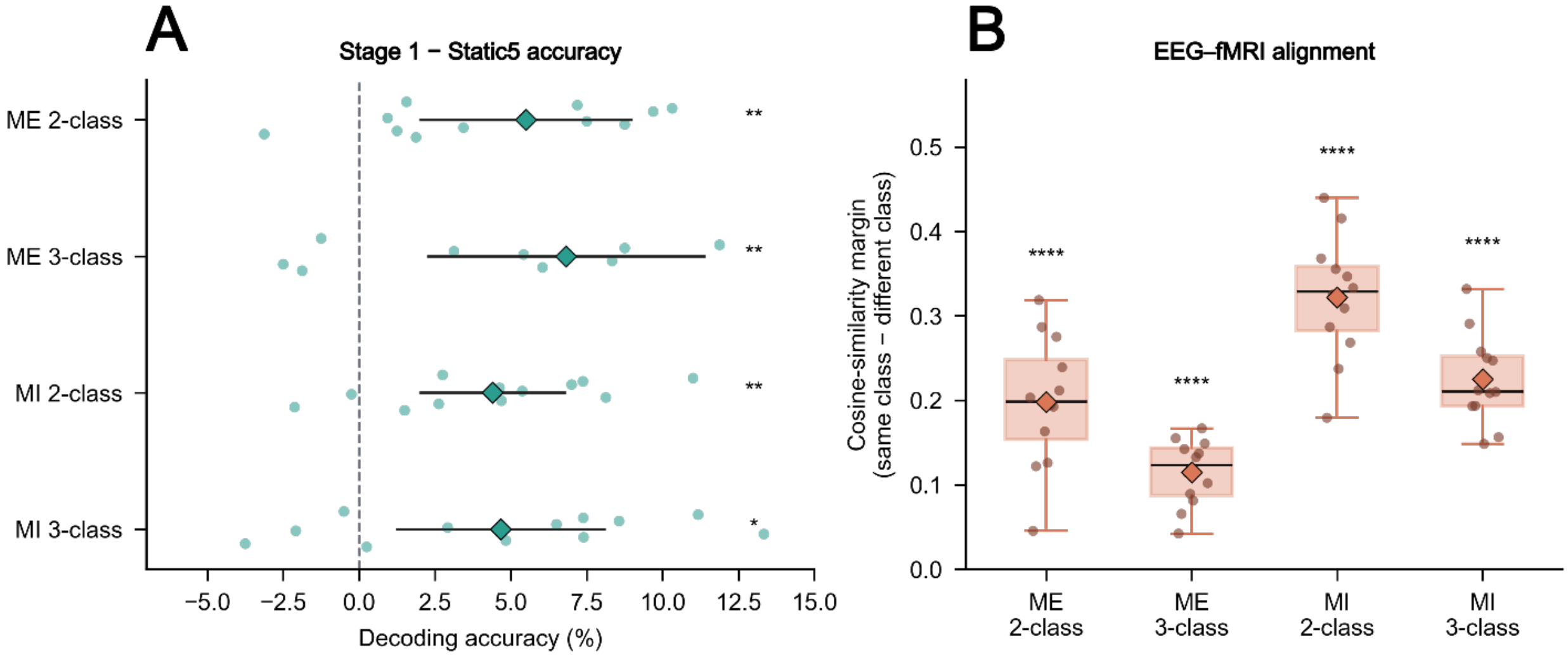

**Fig. 4. Representation-level evidence for the two fMRI-informed stages (EEGNet, full-cohort reference condition)**. **A:** Participant-level accuracy difference between fixed-ridge FRIST-S1 and the fMRI-free Base plus Static5 control. Diamonds show means and horizontal bars show 95% confidence intervals. **B:** Correct-minus-competing fMRI-centroid cosine-similarity margin for held-out EEG embeddings. Boxes show interquartile ranges, center lines show medians, points denote participants, and diamonds show means. Asterisks indicate multiple-comparison corrected statistical significance (*$p_{FDR}$<0.05; **<0.01; ***<0.001; ****<0.0001).

To verify that cross-modal contrastive learning in Stage 2 established a meaningful correspondence between EEG and fMRI representations, we evaluated the prototype similarity margin. Computed on held-out EEG test data using fixed fMRI prototypes, this margin quantifies the difference between an EEG embedding's similarity to the fMRI prototype of the same finger label and its average similarity to the nonmatching finger prototypes. A positive margin therefore indicates that the learned shared space associates the EEG representation more strongly with the matching fMRI class. As shown in Fig. 4B, the prototype margin was positive for every participant under all four decoding conditions. The mean margins were 0.198, 0.115, 0.322, and 0.225 for two-class ME, three-class ME, two-class MI, and three-class MI, respectively. All four margins remained significantly greater than zero after multiple-comparison correction ($p_{\mathrm{FDR}} < 0.0001$). These findings provide direct evidence that Stage 2 consistently learned class-specific cross-modal alignment and complement the incremental-accuracy results shown above.

### 3.4. FRIST Retains EEG Decoding Gains Without Target-Participant fMRI

To determine whether the decoding gain required fMRI from the target EEG participant, we reconstructed both stages of FRIST after excluding that participant's simultaneous fMRI recording. This exclusion had a negligible effect on decoding accuracy (Fig. 5, Supplementary Table S5). Full-cohort versus participant-excluded accuracy after Stage 1 was 75.05% versus 74.77%, 54.73% versus 53.14%, 84.85% versus 85.25%, 67.89% versus 68.43%; final FRIST accuracy was 74.53% versus 73.52%, 56.58% versus 54.70%, 85.63% versus 85.78%, 69.90% versus 70.07%. Seven of the eight full-cohort-versus-participant-excluded comparisons were not significant after correction. The exception was final three-class ME, for which the full-cohort reference was 1.88 points higher.

When both fMRI-derived stages were reconstructed using data exclusively from the other 11 participants, final FRIST accuracies remained 73.52%, 54.70%, 85.78%, and 70.07% for two-class ME, three-class ME, two-class MI, and three-class MI, respectively, corresponding to gains of 6.59, 9.87, 4.99, and 9.14 percentage points over the matched Base. These results indicate that the performance gains of FRIST do not require participant-specific fMRI and can be retained using fMRI-derived references constructed entirely from other members of the cohort.

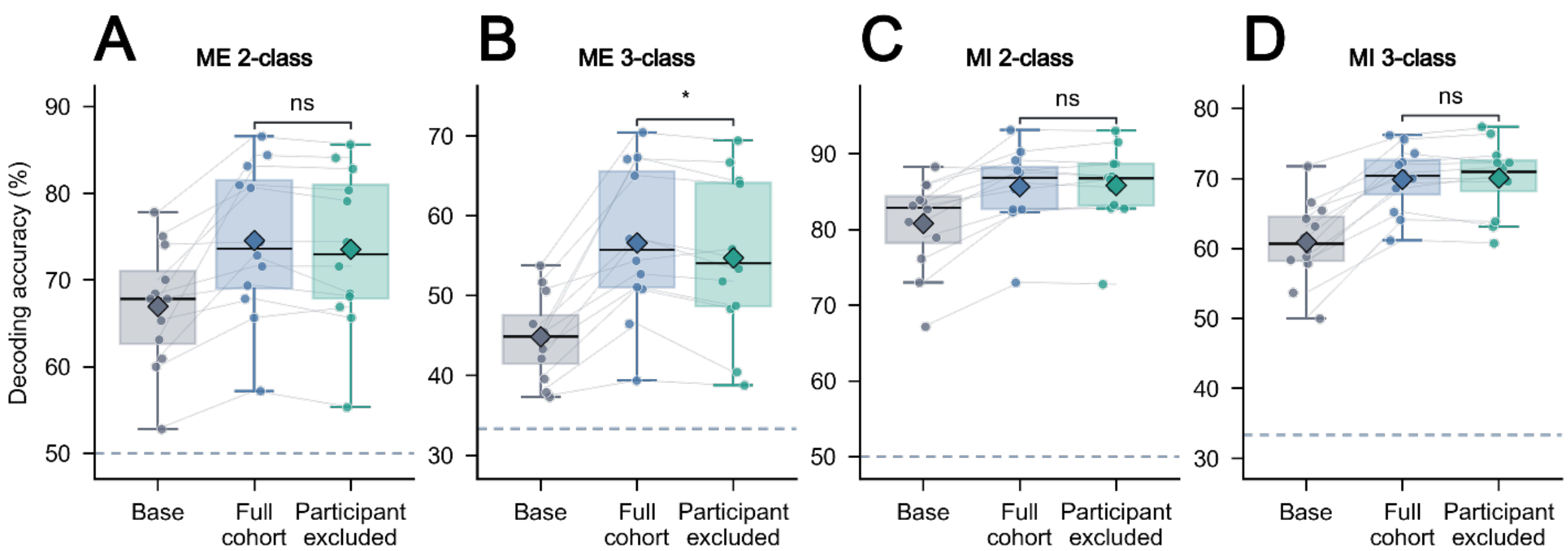

**Fig. 5. Final FRIST accuracy under full-cohort and participant-excluded fMRI-reference construction. A:** Two-class ME. **B**: Three-class ME. **C**: Two-class MI. **D**: Three-class MI. The full-cohort reference used simultaneous EEG-fMRI recordings from all 12 participants, whereas the participant-excluded reference used only the other 11 participants for each target participant. Boxes show interquartile ranges, center lines show medians, whiskers extend

to 1.5 times the interquartile range, points denote participants, diamonds denote means, and gray lines link participant observations. Dashed lines indicate chance. Brackets compare only the two FRIST conditions and summarize two-sided paired tests after Benjamini–Hochberg correction (*p < 0.05; ns, no significance).

### 3.5. FRIST Generalizes Across Multiple EEG Deep Learning Decoders

We then evaluated FRIST across multiple deep-learning-based EEG decoders, including EEGNet (Lawhern et al., 2018), EEG Conformer (Song et al., 2023), a residual dilated temporal convolutional network (TCN) (Bai et al., 2018), and a 1D-CNN-Transformer architecture (Peng et al., 2025), to determine whether its benefit generalized across EEG decoding architectures. FRIST significantly exceeded the corresponding Base model in all 16 matched comparisons (Fig. 6; Supplementary Table S6). Across the four decoding settings, the gains ranged from 4.84 to 11.75 percentage points for EEGNet, 1.97 to 3.90 points for EEG Conformer, 3.52 to 6.36 points for the 1D-CNN–Transformer, and 5.19 to 11.56 points for TCN. EEG Conformer achieved the highest final accuracy in all four settings, reaching 77.68% and 59.67% for two- and three-class ME and 87.40% and 72.54% for two- and three-class MI, respectively. Despite its comparatively strong baseline performance, FRIST significantly improved EEG Conformer in every condition. At the backbone level, mean Base accuracy and mean FRIST gain were strongly negatively correlated ($r = -0.990$, $p = 0.0101$), suggesting that the fMRI-informed representation provided greater complementary benefit to weaker baseline decoders. Overall, the consistent improvements across all four architectures indicate that FRIST is not specific to a particular EEG decoder and can function as a generalizable cross-modal refinement framework.

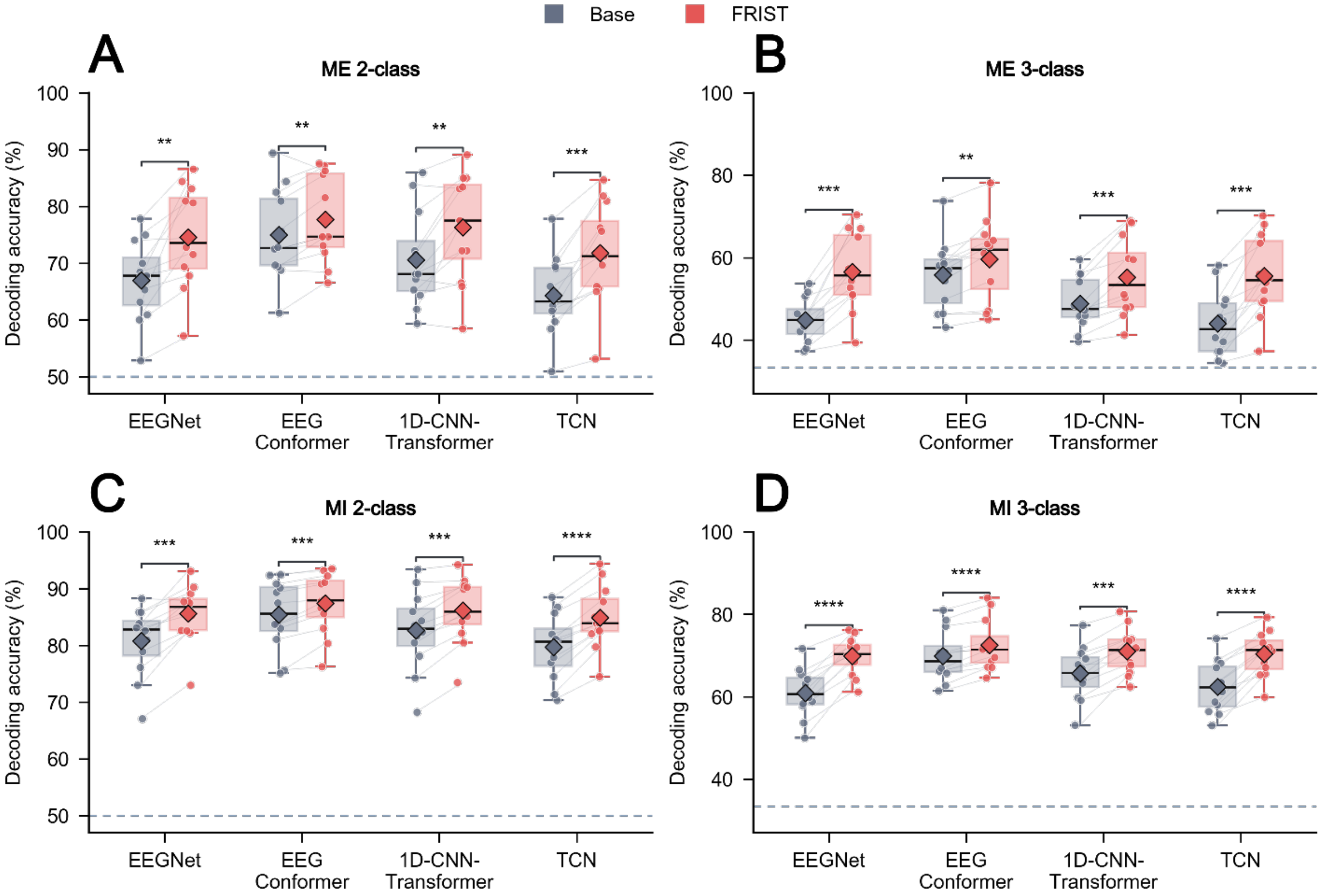

**Fig. 6. Shared-space FRIST performance across four frozen EEG backbones. A**: Two-class ME. **B**: Three-class ME. **C**: Two-class MI. **D**: Three-class MI. Each panel compares participant-level held-out accuracy for the matched EEG-only Base and FRIST using EEGNet, EEG Conformer, 1D-CNN-Transformer, and TCN. Boxes show interquartile ranges, center lines show medians, whiskers extend to 1.5 times the interquartile range, points denote participants, diamonds show means, and dashed lines indicate chance. Brackets show two-sided paired tests after correction within the 16-test final-versus-Base family. Asterisks denote adjusted significance: *$p_{FDR} < 0.05$, **$<0.01$, ***$<0.001$, and ****$<0.0001$; ns denotes no significance.

## 4. DISCUSSION

This study shows that a compact fMRI-informed shared-space training framework can improve individual-finger decoding while keeping the evaluated interface EEG-only. With EEGNet, FRIST accuracy exceeded the matched Base by up to 11.75 percentage points across the four ME and MI settings (Fig. 3). FRIST involves two stages that contributed complementary information: Stage1 provided significant improvements in all four settings through the fMRI informed spectral representation (FISP), whereas Stage2 using residual shared-space alignment and contrastive learning added an additional increment in performance. The largest total gains occurred in the two three-class settings, where finer finger discrimination is most demanding. These results support the central design choice of using fMRI as a model-construction prior rather than as an online signal input.

There is a neurophysiological basis for fMRI-derived class structure to be informative for EEG decoding. Executed and imagined finger movements produce related, although not identical, electrophysiological and hemodynamic responses. EEG captures rapid neural dynamics, including movement-related potentials and mu- and beta-band modulation (Pfurtscheller and Neuper, 2001), whereas fMRI measures slower BOLD responses with greater spatial specificity (Glover, 2011). Simultaneous EEG-fMRI studies have reported spatial correspondence between motor-related alpha/beta modulation and BOLD activity during ME and MI (Bondi et al., 2025; Yuan et al., 2010). At the finger level, multivoxel fMRI studies have further revealed partly overlapping digit representations and multiple action-related maps in the sensorimotor cortex (Diedrichsen et al., 2013; Huber et al., 2020). These findings suggest that finger labels provide a biologically meaningful correspondence between EEG and fMRI representations despite their different temporal scales and signal-generating mechanisms. Our group-level activation analysis supports the spatial premise of the fMRI modality independently of the decoding model (Fig. 2A–E). All fingers engaged in an overlapping sensorimotor network, while the weighted centroids and voxelwise correlations revealed finger-dependent spatial organization. The stronger and more extensive ME response relative to MI further indicates that the two motor conditions were associated with different spatial patterns of BOLD activation. These observations motivate the use of fMRI as a spatial prior.

Motivated by this complementary relationship between EEG and fMRI, previous studies have explored multiple strategies for integrating the two modalities. Simultaneous EEG-fMRI neurofeedback combines both signals during training or feedback to exploit their complementary temporal and spatial information (Perronnet et al., 2017). Other approaches have used fMRI to guide EEG-based motor mapping or spatially weight EEG features according to task-related fMRI activation (Kline et al., 2021; Yang et al., 2022). More recently, cross-modal representation learning has been investigated by relating EEG decoder activity to simultaneously acquired fMRI signals or by jointly pretraining EEG and fMRI representations within a shared embedding space (Iwama et al., 2024; Wei et al., 2025). Collectively, these studies demonstrate that fMRI can provide valuable spatial information for improving EEG analysis. However, many existing approaches rely on simultaneous EEG-fMRI acquisition, participant-specific fMRI data, or synchronized trial pairs, limiting their practicality for routine BCI deployment because of the additional imaging cost, acquisition complexity, and reduced accessibility. FRIST extends these previous EEG-fMRI integration strategies by using fMRI as a source of class-level supervision during model training while preserving a purely EEG-based inference pipeline. Rather than attempting trial-level multimodal decoding or reconstructing one modality from the other, the proposed framework transfers finger-specific spatial organization from simultaneous EEG-fMRI recordings through fMRI informed spectral representation and supervised shared-space alignment.

Stage 1 exploits established coupling between sensorimotor rhythms and BOLD responses to construct a physiologically informed EEG representation (Bondi et al., 2025; Pfurtscheller and Lopes da Silva, 1999; Yuan et al., 2010). Rather than using fMRI as an additional input during decoding, FRIST uses BOLD responses during reference construction to identify time-resolved, band-specific EEG channel projections associated with finger-related hemodynamic activity. These projections are subsequently fixed and applied using EEG alone, allowing fMRI to serve as an offline supervisory signal without imposing a BOLD acquisition requirement at inference. The results in Fig. 3 demonstrate the effectiveness of this strategy: FRIST-S1 improved accuracy over the EEG-only Base model by 4.07–9.91 percentage points across the four decoding conditions. Importantly, the spectral-control analysis in Fig. 4A further distinguishes this improvement from the effect of simply expanding the EEG representation across five frequency bands. The consistent superiority over both the Base and Static5 models support the practical contribution of fMRI guidance to constructing a more informative EEG representation.

Stage 2 builds on the fixed Stage-1 predictor through residual shared-space alignment and supervised contrastive learning. The frozen backbone and FISP features are mapped into a 40-dimensional EEG embedding, while a fixed 40-dimensional fMRI reference bank supplies class-level geometric supervision during training. Rather than replacing Stage 1, a zero-initialized residual head learns only a bounded correction to its scores, so the initial Stage-2 prediction exactly recovers the Stage-1 prediction (Friedman, 2001; He et al., 2016). Training combines classification and residual-direction objectives with fMRI class-mean alignment, a fixed fMRI LDA objective, and a one-directional cross-modal adaptation of supervised contrastive learning. The contrastive term uses EEG embeddings as anchors, same-finger fMRI examples as positives, and different-finger examples as negatives, thereby transferring class structure without requiring paired EEG–fMRI trials (Khosla et al., 2020). A compact affine residual calibrator is subsequently fitted using validation data only. During inference, prediction uses the frozen EEG paths, learned residual head, and validation-fitted calibrator; neither the fMRI bank, fMRI centroids, nor fMRI-derived scores enter the decoding pathway. Relative to FRIST-S1, Stage 2 changed accuracy by −0.52, +1.85, +0.77, and +2.00 percentage points across the four respective settings. The additional gains were significant for three-class ME and for both MI settings, whereas the change in two-class ME was not significant. The analysis-only correct-centroid cosine-similarity margin was positive for every participant in every setting (Fig. 4B; Supplementary Table S7). Mean ± SD margins were 0.198 ± 0.077, 0.115 ± 0.039, 0.322 ± 0.073, and 0.225 ± 0.053, and all four remained significantly greater than zero after correction. These findings indicate that held-out EEG embeddings consistently reflected the class organization of the fMRI reference, although the fMRI centroids themselves were not used for classification.

Within the broader BCI literature, FRIST represents a step toward improving noninvasive individual-finger decoding through cross-modal fMRI guidance. Invasive recordings have enabled individual-finger decoding and prosthetic-finger control (Hotson et al., 2016; Kubánek et al., 2009; Willsey et al., 2025), and intracortical BCIs have been coupled with neuromuscular stimulation to restore functional upper-limb movement (Bouton et al., 2016). Although these systems provide high spatial specificity and fine motor control, their broader use is constrained by the need for surgical implantation. EEG-based BCIs are noninvasive and more accessible, but finger-specific cortical activity is spatially blurred by volume conduction and mixed at the scalp. Consequently, many EEG motor BCIs have focused on coarse movement classes. Recent studies have nevertheless demonstrated that scalp EEG contains sufficient information to discriminate individual fingers offline and, more recently, to enable real-time robotic finger control (Ding et al., 2025; Lee et al., 2022; Liao et al., 2014). Despite these advances, decoding accuracy remains substantially lower than that achieved with invasive approaches, limiting the robustness required for practical finger-based BCI applications.

The present work addresses this need by demonstrating that spatial information derived from fMRI can improve EEG-only finger decoding without changing the modality required during deployment. The chronological session-held-out protocol further provides a more realistic assessment than random trial splitting: each test session was decoded using only EEG data collected in preceding sessions for model development. The EEG pipeline also operates on 1-s windows, providing a computational basis for low-latency decoding. Nevertheless, the present evaluation remained offline, and prospective closed-loop experiments are still required to determine whether the observed improvements translate to real-time control under feedback, nonstationarity, and user adaptation.

The participant-excluded analysis further demonstrated the practical potential of the proposed framework by showing that fMRI guidance remained effective even when the target participant's fMRI data were excluded from the training set (Fig. 5). The framework leveraged finger-specific representations learned from a population-level EEG-fMRI dataset to improve EEG decoding for previously unseen participants. This finding suggests that a previously acquired EEG-fMRI dataset may be sufficient to provide meaningful supervision for new EEG users, making the framework considerably more scalable for practical deployment without the need for time-consuming and costly multimodal data acquisition. Although future studies with larger cohorts will be needed to further characterize the generalizability of population-level fMRI representations across broader populations, these results provide initial evidence that transferable finger-specific neural representations can be exploited to enhance subject-specific EEG decoding.

In addition, the improvements were consistently observed across multiple deep-learning EEG decoders, including EEGNet (Lawhern et al., 2018), EEG Conformer (Song et al., 2023), TCN (Bai et al., 2018), and 1D-CNN–Transformer (Peng et al., 2025). Final FRIST significantly exceeded the corresponding Base model in all 16 matched comparisons (Fig. 6; Supplementary Table S6). This suggests that the proposed framework is not tied to a particular network architecture but instead functions as a general representation-learning strategy that can complement diverse EEG decoders. Because the framework operates by enriching the learned EEG feature space rather than modifying a specific classifier design, it represents a flexible cross-modal learning paradigm that may continue to improve alongside advances in EEG deep-learning architectures. The exploratory inverse correlation between baseline performance and gain is consistent with stronger encoders already recovering part of the information supplied by the fMRI prior.

Several limitations should be considered. First, the evaluation was limited to two- and three-finger decoding tasks. Extending the proposed framework to all five fingers and more complex finger motions will be important for assessing its scalability to higher-dimensional BCI control. Second, although FRIST consistently improved EEG decoding, the size of the EEG-fMRI dataset remained relatively modest. A larger and more diverse cohort may enable the learning of more discriminative and generalizable fMRI-derived finger representations, potentially leading to further improvements in EEG decoding performance. Third, the present evaluation was performed offline using held-out sessions. Although the chronological held-session protocol approximates an online deployment scenario by preventing future-session leakage, it does not fully capture the nonstationarity, feedback effects, and user adaptation that occur during real-time BCI control. Prospective online testing will therefore be important to determine whether fMRI-guided representation learning improves real-time individual-finger BCI performance.

Future work should focus on both prospective online validation and broader neuroscientific interpretation. First, additional online BCI experiments should be performed to test whether the FRIST improves real-time individual-finger control under closed-loop conditions. These experiments will be important for evaluating robustness to nonstationary EEG signals, online feedback, and user adaptation. Second, the current two- and three-finger settings should be extended to all five fingers as more online data become available and examine whether fMRI-derived spatial priors remain beneficial for more complex finger-level decoding. Third, the neural relationship between EEG and fMRI representations during fine motor execution and imagery should be further investigated. By analyzing how EEG embeddings align with fMRI-derived finger prototypes, future work may help clarify how electrophysiological dynamics and hemodynamic spatial patterns jointly encode individual-finger information. In this way, fMRI-guided EEG modeling may serve not only as a decoding strategy, but also as a computational framework for studying cross-modal neural representations of human fine motor control.

## 5. CONCLUSION

We have introduced an fMRI representation informed shared-space training framework for improving EEG-only individual-finger decoding. We demonstrate that FRIST improves EEG-only individual-finger decoding by transferring EEG–BOLD spectral coupling and finger-discriminative fMRI structure from a separate simultaneous EEG–fMRI reference dataset. Its fixed fMRI-informed spectral projection and residual shared-space alignment together with contrastive learning provide complementary benefits across two- and three-class ME and MI tasks, while the backbone analysis show robust performance enhancement across state-of-the-art EEG decoders. Most of these gains were retained in the group of participates studied when the target participant's fMRI was excluded, supporting the potential reuse of population-level reference data for new EEG users. Because inference requires only EEG and the learned model parameters, the present results suggest that FRIST preserves the accessibility of EEG-based BCIs while benefiting from fMRI during model construction. Although the present evaluation was performed offline using chronological held-out sessions, the results support further prospective evaluation of fMRI-guided representation learning in closed-loop EEG finger-level BCI systems.

## Acknowledgments

We thank Dr. Elena Bondi for useful discussions on EEG-fMRI experiments, and Yisha Zhang for technical assistance in data collection. This work was supported in part by National Institutes of Health grants NS124564 (BH), NS131069 (BH), NS127849 (BH), and NS096761(BH). Y.D. was supported in part by a Carnegie Mellon University's Center for Machine Learning and Health Fellowship. J.K. was supported in part by the NSF Graduate Research Fellowship Program under Grant No. DGE2140739. J.K. and M.K. were partially supported by National Institutes of Health T32 training grant EB029365 (PI: BH).

## Data Availability

All data supporting the findings of this study are available within the article and its supplementary files. The EEG and fMRI data in all subjects used in this study will be made available in a public repository upon paper acceptance.

## Code Availability

Customized codes used in this study will be available on GitHub upon paper acceptance.